\documentclass[sigconf]{acmart}
\usepackage{multirow}
\usepackage{tikz-dependency}
\usepackage{balance}
\def\BibTeX{{\rm B\kern-.05em{\sc i\kern-.025em b}\kern-.08emT\kern-.1667em\lower.7ex\hbox{E}\kern-.125emX}}
    
\copyrightyear{2019}
\acmYear{2019}
\setcopyright{acmcopyright}
\acmConference[CIKM '19] {The 28th ACM International Conference on Information and Knowledge Management}{November 3--7, 2019}{Beijing, China}
\acmBooktitle{The 28th ACM International Conference on Information and Knowledge Management (CIKM'19), November 3--7, 2019, Beijing, China}
\acmPrice{15.00}
\acmDOI{10.1145/3357384.3358142}
\acmISBN{978-1-4503-6976-3/19/11}

\begin{document}

\fancyhead{}
  % do not delete this code.

% The "title" command has an optional parameter, allowing the author to define a "short title" to be used in page headers.
\title{Aspect and Opinion Aware Abstractive Review Summarization with Reinforced Hard Typed Decoder}
%\title{Aspect and Opinion Aware Review Headline Generation with Reinforced Hard Typed Decoder}
\author{Yufei Tian}
\affiliation{%
  \institution{Tsinghua University}
  %\city{Beijing}
  %\country{China}
}
\email{tyf16@mails.tsinghua.edu.cn}

\author{Jianfei Yu}
\authornote{Corresponding author.}
\affiliation{%
  \institution{Singapore Management University}}
\email{jfyu@smu.edu.sg}

\author{Jing Jiang}
\affiliation{%
  \institution{Singapore Management University}
}
\email{jingjiang@smu.edu.sg}

%
% The abstract is a short summary of the work to be presented in the article.
\begin{abstract}
In this paper, we study abstractive review summarization.
Observing that review summaries often consist of aspect words, opinion words and context words, we propose a two-stage reinforcement learning approach, which first predicts the output word type from the three types, and then leverages the predicted word type to generate the final word distribution.
Experimental results on two Amazon product review datasets demonstrate that our method can consistently outperform several strong baseline approaches based on ROUGE scores.
\end{abstract}

%
% The code below is generated by the tool at http://dl.acm.org/ccs.cfm.
% Please copy and paste the code instead of the example below.
%

%\ccsdesc[500]{Computer systems organization~Embedded systems}
%\ccsdesc[300]{Computer systems organization~Redundancy}
%\ccsdesc{Computer systems organization~Robotics}
%\ccsdesc[100]{Networks~Network reliability}

%
% Keywords. The author(s) should pick words that accurately describe the work being
% presented. Separate the keywords with commas.
%
% This command processes the author and affiliation and title information and builds
% the first part of the formatted document.
\maketitle

\section{Introduction}
\label{sec:intro}

Reviews posted on e-commerce platforms by online shoppers are valuable resources for businesses to improve product quality and keep track of customer preferences.  
However, given the large amount of product reviews in real scenarios, it is impractical to manually read through each review, especially when they are lengthy and have low readability. 
Therefore, it is crucial to design a robust model to automatically generate concise and readable summaries for product reviews, which is often referred to as \textit{review summarization}~\cite{ganesan:coling2010}.
%In this paper, we focus on the implementation of new end-to-end models to generate abstractive summaries. 

In the literature, existing approaches to \textit{review summarization} generally belong to two groups: extractive summarization and abstractive summarization.
The former line of work focuses on selecting several informative sentences or phrases from a set of reviews of a product~\cite{wang:naacl2016,angelidis:emnlp2018}.
The latter centers on generating a short but meaningful summary for either a single review of a product~\cite{yang2018aspect,yang:cikm2018} or multiple reviews of a product~\cite{ganesan:coling2010,gerani:emnlp2014}.
In this paper, following Yang et al.~\cite{yang2018aspect}, we aim to develop an effective model that can generate a concise summary for a single input review.

\begin{figure}[!tp]
	\setlength{\abovecaptionskip}{0cm}
	\setlength{\belowcaptionskip}{-0.5cm}
    \centering
    \includegraphics[width = 0.8\linewidth]{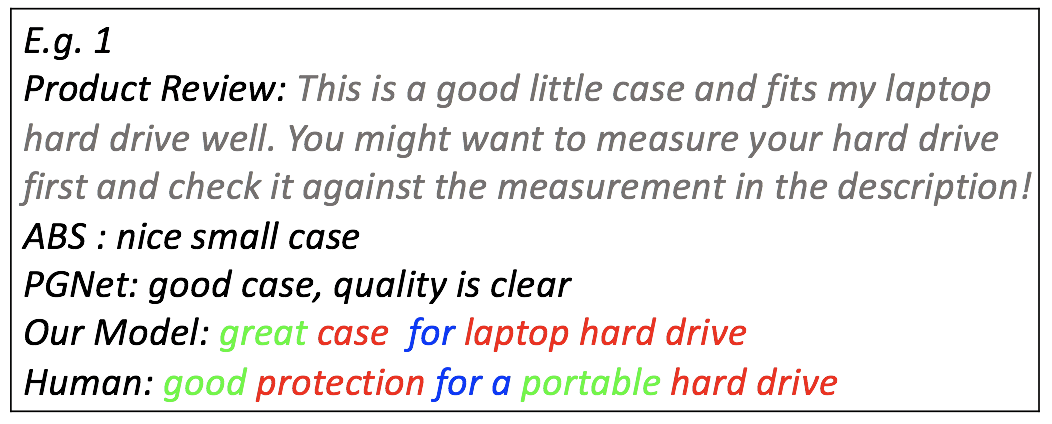}
    \caption{Current models tend to output general and less meaningful summaries.
}
    \label{fig:1}
    \setlength{\abovecaptionskip}{0cm}
	\setlength{\belowcaptionskip}{-0.5cm}
\end{figure}

As with any abstractive summarization task, current representative encoder-decoder frameworks including the well-known sequence-to-sequence~(\textit{Seq2Seq}) model~\cite{nallapati2016abstractive} and the advanced pointer-generator network~(\textit{PGNet})~\cite{gu:acl2016,see2017get} %,paulus2017deep}
are natural to be adopted.
More recently, observing that some under-represented aspect and opinion words tend to be ignored by \textit{Seq2Seq} and \textit{PGNet}, \citet{yang2018aspect} proposed a multi-factor attention network based on \textit{PGNet}, which forces their model to focus more on aspect and opinion words in their modified attention mechanism.
%Previous study on abstractive summarization can be seen in \cite{nallapati2016abstractive,paulus2017deep,see2017get}, in which a seq2seq framework  is employed to generate summaries on the CNN/Daily Mail. \citet{see2017get} used a pointer-generator network that can copy words from the source text via pointing. 
%However, these methods share two common limitations:  they tend to {\bf1)} output generic responses like 'great product for the price'  and {\bf2)} cannot identify objects referred to as 'it' or 'this' throughout the review.
However, since all these approaches assume to generate words from the same vocabulary at each decoding step, they tend to produce generic summaries with high-frequency phrases as in Fig.~\ref{fig:1}, which often fail to include those less frequent aspect or opinion words that are also essential to review summaries.
As illustrated in Fig.\ref{fig:1}, an informative review summary written by human should be a natural composition of \textit{\color{red}aspect words}, \textit{\color{green}opinion words}, and \textit{\color{blue}context words}, where \textit{\color{red}aspect words} and \textit{\color{green}opinion words} indicate the product information and users' opinions respectively, and \textit{\color{blue}context words} are used to make the summary coherent. %\jfcomment{Include human summary in Fig.\ref{fig:1}.}

%The idea presented by \citet{wang2018learning} to control word types (from interrogatives, topic words and ordinary words) in the context of question answering can be used as reference, but their model is based on seq2seq, which fails to copy OOV words from the source text. Also, their Hard Typed Decoder applied Gumbel-Softmax \citep{jang2016categorical} as an differentiable surrogate of argmax, which might lead to noises or suboptimal solutions.

Motivated by this, we borrow the idea from a state-of-the-art dialogue question generation model~\cite{wang2018learning}, and aim to explicitly control word types when generating the review summary.
Specifically, we first classify all the vocabulary words into three types: \textit{\color{red}aspect words}, \textit{\color{green}opinion words}, and \textit{\color{blue}context words}.
Based on this, we 
%adapt the Hard Typed Decoder~(\textit{HTD}) proposed in~\cite{wang2018learning} to \textit{PGNet}, and 
propose a two-stage Reinforced Hard Typed Decoder~(\textit{RHTD}), which first explicitly predicts the output word type, followed by generating the final word distribution based on the predicted type at each decoding position.
Due to the discrete choice of word types, the gradient over the first stage becomes non-differentiable.
To jointly optimize the two stages, instead of simply following \citet{wang2018learning} to use  Gumbel-Softmax~\cite{jang2016categorical} as a differentiable approximation, we adopt a widely used policy gradient algorithm REINFORCE~\cite{williams1992simple} to assign an explicit feedback reward to the predicted word type.%, which is determined by whether the type of the final generated word is the same as the reference word type.
%The second stage is optimized with the standard maximum-likelihood estimation objective function, while the first stage is trained using REINFORCE~\cite{williams1992simple} with a feedback reward from the second stage, which is determined by whether the final generated word has the same type as the referenced word type.
%Specifically, instead of using \textit{Seq2Seq} as the base model as in~\newcite{wang2018learning}, we build our model on top of the more advanced \textit{PGNet}, where both the generation distribution and the pointer distribution are dependent on the word type distribution.
%Moreover, different from~\newcite{wang2018learning} that simply uses Gumbel-Softmax~\cite{jang2016categorical} 
%We also applied policy gradient \citep{sutton2000policy} to ... .

We carry out experiments on two Amazon product review datasets.
Automatic evaluation based on ROUGE scores~\cite{lin:2004:ACLsummarization} demonstrates that \textit{RHTD} outperforms several highly competitive baseline approaches. %including our modified \textit{HTD} model.
Further analysis verifies that \textit{RHTD} can indeed produce more informative summaries.

\section{Methodology}
\label{sec:method}
In this section, we first formulate our task.
We then introduce our aspect and opinion words extraction approach, and review Pointer-Generator Network.
Finally, we describe a modified Hard Typed Decoder model, followed by our reinforcement learning method.

\subsection{Task Definition}
We are given a set of user product reviews $\mathcal{D}$, and each product review $X \in \mathcal{D}$ is associated with a short summary $Y$.
The task of abstractive review summarization can be formalized as follows: given a source product review with $m$ words $X = x_1, x_2, ..., x_m$, the system should generate a concise and informative target summary with $n$ words $Y^* = y_1, y_2, ..., y_n$ that captures the salient points, formally as
$ Y^{*} = \mathop{argmax}_{Y} P(Y|X) $.

\subsection{Aspect and Opinion Words Extraction}
\label{DP:method}
%Figure 1 shows that a well-informed summarization system should be aware of word categories: aspect, opinion and general. 
As mentioned before, we assume that a well-informed review summary should consist of three types of words: aspect, opinion, and context words.
However, due to the labor intensive
nature of human annotation, it is almost impossible to manually collect all the aspect and opinion words.
Therefore, we employ the well-known unsupervised extraction method, Double Propagation \cite{qiu2011opinion}, to automatically extract all the aspect and opinion words in each domain. 

\begin{table}[!tp]
\centering
\small
\setlength{\abovecaptionskip}{0cm}
\setlength{\belowcaptionskip}{0cm}
\begin{tabular}{|c|c|c|}
\hline
\textbf{Rules}   & \hspace{0.5em}\textbf{Relations (PoS Tags)}       & \hspace{0.3em} \textbf{Examples}     \\\hline
\begin{dependency}[theme = simple]
   \begin{deptext}[column sep=1em]
      R1 \\
   \end{deptext}
\end{dependency} &
\begin{dependency}[theme = simple]
   \begin{deptext}[column sep=1em]
      AP (\textbf{NN}) \& \hspace{0.3em}AP (\textbf{NN}) \\
   \end{deptext}
   \depedge[arc angle=25]{1}{2}{nn}
\end{dependency} &
\begin{dependency}[theme = simple]
   \begin{deptext}[column sep=1em]
     The \& Mac \& OS \& is excellent. \\
   \end{deptext}
   \depedge[arc angle=20]{2}{3}{nn}
\end{dependency}    \\ \hline
\begin{dependency}[theme = simple]
   \begin{deptext}[column sep=1em]
      R2 \\
   \end{deptext}
\end{dependency} &
\begin{dependency}[theme = simple]
   \begin{deptext}[column sep=1em]
      OP (\textbf{JJ}) \&\hspace{0.8em} OP (\textbf{JJ}) \\
   \end{deptext}
   \depedge[arc angle=25]{1}{2}{conj}
\end{dependency} &
\begin{dependency}[theme = simple]
   \begin{deptext}[column sep=1em]
     It is \& light \& and \& portable. \\
   \end{deptext}
   \depedge[arc angle=25]{4}{2}{conj}
\end{dependency} \\ \hline
\begin{dependency}[theme = simple]
   \begin{deptext}[column sep=1em]
      R3 \\
   \end{deptext}
\end{dependency} &
\begin{dependency}[theme = simple]
   \begin{deptext}[column sep=1em]
      AP (\textbf{NN}) \& \hspace{0.3em} OP (\textbf{JJ}) \\
   \end{deptext}
   \depedge[arc angle=25]{1}{2}{nsubj}
\end{dependency} &
\begin{dependency}[theme = simple]
   \begin{deptext}[column sep=1em]
     The \& speed \& is \& incredible. \\
   \end{deptext}
   \depedge[arc angle=25]{2}{4}{nsubj}
\end{dependency}    \\ \hline
\begin{dependency}[theme = simple]
   \begin{deptext}[column sep=1em]
      R4 \\
   \end{deptext}
\end{dependency}  &
\begin{dependency}[theme = simple]
   \begin{deptext}[column sep=1em]
      OP (\textbf{JJ}) \& \quad AP(\textbf{NN}) \\
   \end{deptext}
   \depedge[arc angle=25]{1}{2}{amod}
\end{dependency}  &
\begin{dependency}[theme = simple]
   \begin{deptext}[column sep=1em]
     iPhone has \& great \& design. \\
   \end{deptext}
   \depedge[arc angle=20]{2}{3}{amod}
\end{dependency}  \\\hline
\end{tabular}
\caption{\footnotesize Rules for extracting aspect words (\textbf{AP}) and opinion words (\textbf{OP}). \textbf{NN} and \textbf{JJ} respectively denote two sets of Part-of-Speech Tags (PoS Tags), i.e., \{NN, NNS\} and \{JJ, JJS and JJR\}. \textit{conj}, \textit{nn}, \textit{amod}, and \textit{nsubj} are dependency relations.}
\label{tab:synrel}
\vspace{-0.5cm}
\end{table}

Following Qiu et al.~\cite{qiu2011opinion}, we leverage four syntactic rules in Table~\ref{tab:synrel} to identify potential aspect and opinion words.
%, which are based on four dependency relations: \textit{conj}, \textit{nn}, \textit{amod}, and \textit{nsubj},  to identify potential aspect and opinion words.
Specifically, we first utilize a sentiment lexicon\footnote{https://www.cs.uic.edu/$\scriptsize{\sim}$liub/FBS/sentiment-analysis.html\#lexicon} to extract all the opinion words occurring in source product reviews of $\mathcal{D}$, and then expand the opinion word list based on \textit{R2} in Table~\ref{tab:synrel}. 
Given the extracted opinion words, we further use \textit{R3} and \textit{R4} to extract aspect words from $\mathcal{D}$.
For example, in \textit{R3}, since \textit{incredible} is detected as an opinion word and the subject of \textit{incredible} is usually aspect words, we can employ this rule to detect that its subject \textit{speed} is an aspect word.
Next, the aspect word list is also expanded based on \textit{R1}.
Finally, we can make use of the above four rules to iteratively expand the aspect and opinion word lists.
Based on the identified aspect and opinion words, let us use $A$, $O$, and $C$ to respectively denote the three word types \{aspects, opinions, context words\}, $V$ the whole vocabulary.
%and $V_A$, $V_O$, $V_C$, $V$ the aspect, opinon, context and whole vocabulary, respectively.

\subsection{Pointer-Generator Network (PGNet)}
\label{pgnet}
Since \textit{PGNet} is essentially a combination of \textit{Seq2Seq}~\cite{nallapati2016abstractive} and a pointer network~\cite{vinyals:nips2015} and has been shown to outperform \textit{Seq2Seq} in many generation tasks~\cite{gu:acl2016,see2017get}, we adopt it as our base model.
First, let us introduce the necessary notation for \textit{Seq2Seq}.
We use $s_t$ to denote the decoder state at time step $t$, $a^t_k$ the attention weight over each encoder hidden state $h_k$, and $h^{\star}_t$ the weighted sum of encoder hidden states.
To generate the word distribution over $V$ at time step $t$, $h^{\star}_t$ and $s_t$ are concatenated together by feeding them to a linear function:
\begin{equation}
\label{vocab}
%\notag
P_{\text{vocab}}(w_t) = \text{softmax}(W^T[s_t,h^{\star}_t]+b),
\end{equation}
where $W$ and $b$ are learnable parameters.

In \textit{PGNet}, a generation probability $p_{\text{gen}} \in [0,1]$ is introduced to control whether to generate a word from $V$ or copy words from the input sequence $X$ via pointer network at time step $t$:
\begin{equation}
    p_{\text{gen}} = \sigma(w^T_{h}h^{\star}_t + w^T_ss_t + w^T_xx_t + b_{ptr})
\end{equation}
where $w_{h}$, $w_s$, $w_x$ and $b_{ptr}$ are parameters to be learned.
The final probability distribution over the extended vocabulary is: 
\begin{equation}
    P_{\text{vocab}}(w_t) = p_{\text{gen}}P_{\text{vocab}}(w_t)+ (1 {-} p_{\text{gen}})\sum\limits_{k: w_k= w_t}a^t_k,
    \label{pointer}
\end{equation}
where the first and second terms are respectively referring to the generation distribution in Eq. (\ref{vocab}) and the distribution over the input sequence $X$ by sampling from the attention distribution $a^t$ for the encoder hidden states.

\subsection{Proposed Approach}
\label{PA}
Recall that to help our model pay more attention to aspect and opinion words to generate more informative summary, we propose to first explicitly control the type of the output word, followed by generating the word distribution based on the predicted type at each decoding step.
We formulate this process as follows:
\begin{align}
\label{eq3}  c^{*}_t = \mathop{\arg\max}_{c_i} P(tp_{w_t} & = c_i| w_{<t},X), i \in \{0,1,2\},\\
   P_{\text{vocab}(c^{*}_t)}(w_t) & = P(w_t|tp_{w_t} = c^{*}_t),
          % & = \text{softmax}(W_{c^{*}}^T[s_t,h^{*}_t]+b).
\end{align}
where $c_i$ is one of the three word types \{$A$, $O$, $C$\} and $tp_{w_t}$ denotes the word type at time step $t$. 
Note that we split all the words in $V$ and $X$ into aspect, opinion, and context words respectively, and $P_{\text{vocab}(c^{*}_t)}(w_t)$ is a type-specific word distribution.

\begin{figure}[!tp]
%\label{fg2}
\centering
\includegraphics[scale=0.29]{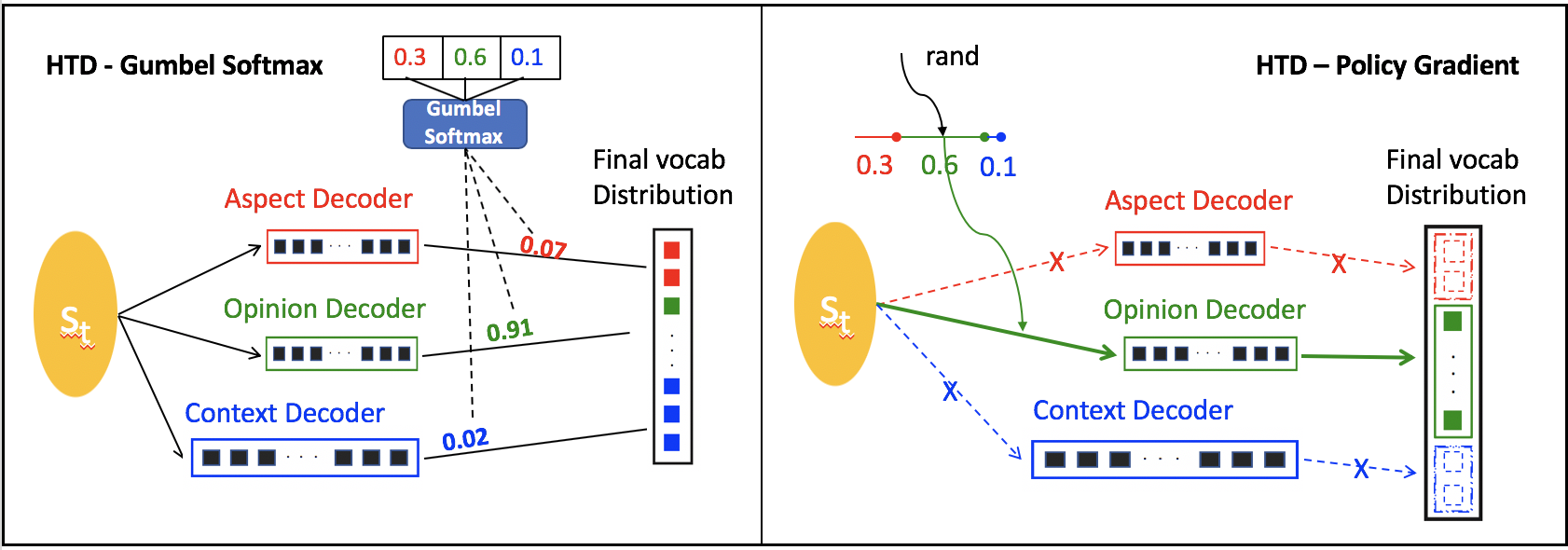}

\vspace*{-4mm}
\caption{Overview of Hard Typed Decoder (left) and Our Reinforced Hard Typed Decoder (right).}
\vspace{-3mm}
\label{decoder_framework}

\end{figure}

However, the choice of word type (i.e., $\text{argmax}$) is discrete and non-differentiable.
Therefore, we propose the following two solutions to tackle this problem. 

\vspace{0.5em}
\noindent\textbf{2.4.1 \hspace{0.3em} Hard Typed Decoder}

As shown in Fig.~\ref{decoder_framework}, we first borrow the idea from the Hard Typed Decoder (\textit{HTD}) proposed by \citet{wang2018learning}, and use Gumbel-Softmax (GS)~\cite{jang2016categorical} as a differentiable surrogate.
As the \textit{HTD} model employed by \citet{wang2018learning} is simply based on \textit{Seq2Seq}, here we adapt it to \textit{PGNet} with some modifications.
Specifically, to approximate $P_{\text{vocab}}(w_t)$, we introduce:
\begin{equation}
    P^{'}_{\text{vocab}}(w_t) = P(w_t|tp_{w_t} = c_i) \cdot \textbf{\emph{m}}(w_t), i \in \{0,1,2\},
    \label{P_w}
\end{equation}
where 
\begin{align}
    \textbf{\emph{m}}(w_t) & = \textbf{GS}\big(P(tp_{w_t} = c_i| w_{<t},X)\big), i \in \{0,1,2\}, \\
    & \textbf{GS}(X_i) = \frac{e^{(log(X_i)+g_i)/ \tau}}{\sum_{j = 0}^{2}e^{(log(X_j)+g_j)/ \tau}}.
\end{align}
Note that $g_i$ are i.i.d samples drawn from the Gumble distribution, and $\tau$ is a hyper-parameter to control the smoothness of the distribution.
The closer constant $\tau$ is to 0, the similar Eq.(\ref{P_w}) is to $\text{argmax}$. We set $\tau$ to 1 to make GS smoother than $\text{argmax}$ but can also exhibit the \textbf{hard} characteristics.
 
Similarly, we modify the copy distribution $a^t_k$ over the input sequence $X$ to be $ \beta^{t}_k = a^{t}_k \cdot \textbf{\emph{m}}(w_t)$, and the final generation probability distribution can be calculated as follows:
\begin{align}
P(w_t) =  p_{\text{gen}}P^{'}_{\text{vocab}}(w_t) + (1- p_{\text{gen}})\sum\limits_{k: w_k= w_t}\beta^{t}_k.
\end{align}

Finally, the loss function for \textit{HTD} is essentially a combination of copy, generation, and type loss:
\begin{align}
    & \mathcal{J}_{h} = \sum_{t} -\big(\log P(w_t) + \lambda \log P(tp_{w_t}| w_{<t},X)\big)
    \label{equation: lambda}
\end{align}
where $\lambda$ is a hyper-parameter.

\vspace{0.5em}
\noindent\textbf{2.4.2 \hspace{0.3em} Reinforced Hard Typed Decoder}

\textbf{Motivation:} Although the above \textit{HTD} model can eliminate the non-differentiable gradient issue, it mainly suffers from the following problem.
Since the modified GS distribution is much sharper than Softmax, it may lead to severe error propagation to the following word generation process if the original Softmax distribution significantly deviates from the reference type distribution.
Inspired by this, we propose to jointly train the two stages with REINFORCE algorithm~\cite{williams1992simple}, which can largely eliminate the error propagation issue by sampling a word type based on the original Softmax distribution.

Specifically, we first initialize all the model parameters with a well-trained \textit{HTD} model.
Given an input review $X$ and its generated word $w_{<t}$ before time step $t$, we first calculate the type distribution $P(tp_{w_t}|w_{<t}, X)$ in Eq.(\ref{eq3}), followed by sampling a word type $c(\bar{w_t})$ at time step $t$, where $\bar{w_t}$ denotes the output word from the second stage.
Then, the gradient for the second stage is as below:
\begin{align}
    & \nabla_{\phi_2}\mathcal{J}_{2}(\phi_2) = \nabla_{\phi_2}-\log P(w_t)
    \label{equation: lambda}
\end{align}
Next, the rewards for training the first stage is calculated as follows:
\begin{equation}
v_t = \left\{
             \begin{array}{lr}
             0.3, \quad c(\bar{w_t}) \ne c(w^{*}_t) &  \\
             1.0, \quad c(\bar{w_t}) = c(w^{*}_t) & 
             \end{array}
\right.
\label{eq:reward_factor}
\end{equation}
where $c(w^{*}_t)$ is the reference word type.
The gradient for the first stage is then computed based the policy gradient theorem~\cite{williams1992simple}:
\begin{align}
\notag
    & \nabla_{\phi_1}\mathcal{J}_{1}(\phi_1) = \mathbb{E}[v_t \cdot \nabla_{\phi_1}(-\log P(tp_{w_t}|w_{<t},X))]
    \label{equation: lambda}
\end{align}
where the sampling approach is used to estimate the expected reward.
We repeat the above iterative training process until convergence.

\section{Experiments}
\label{sec:exp}

%In this section, we evaluate performance of different approaches on review summarization.
\subsection{Experiment Settings}
\par  \textbf{Datasets:} We evaluate our model on Amazon reviews dataset\footnote{http://jmcauley.ucsd.edu/data/amazon/.}, 
%which covers a variety of products such as books, electronics, video games and music.
and select two domains from the raw dataset to conduct our experiments, which are \texttt{Healthcare} and \texttt{Electronics}. 

\textbf{Pre-processing Details:}
For both datasets, we filter out review-summary pairs that are too long/short to expedite training and testing, and obtained 48,495 and 187,143 valid review-summary pairs.
We then randomly split them into training (70\%), development (10\%) and test sets (20\%).
Next, as introduced in Section~\ref{DP:method}, we applied Double Propagation method~\cite{qiu2011opinion} on the training set to extract 3,104 aspect words and 2,118 opinion words for \texttt{Healthcare} domain.
The number for \texttt{Electronics} domain is 14,305 and 11,232.

\textbf{Parameter Settings:}
For all the experiments, we set the word embedding size $e$ to be 128, and initialize the word embedding matrix $\mathbf{E}$ using pre-trained word embeddings based on Glove\footnote{https://nlp.stanford.edu/projects/glove/.}, which will be fixed during the training process.
The hidden dimension $d$ and the number of LSTM layers in both datasets are set to be 128 and 1.
During training, we adopt Adagrad \citep{duchi2011adaptive} with learning rate 0.05. 
Note that we initialize the parameters in \emph{RHTD} with a pre-trained \emph{HTD} model.

\textbf{Evaluation Metrics:} 
Following many previous studies on abstractive summarization, we choose ROUGE-1, 2, L~\citep{lin:2004:ACLsummarization} to automatically quantify how well a model fits the data.

\subsection{Main Results}

%\vspace{-0.6cm}
\begin{table}[H]
\begin{tabular}{|l|c|c|c|c|c|c|}
\hline
{\color[HTML]{000000} }                                 & \multicolumn{3}{c|}{{\color[HTML]{000000} \textbf{Healthcare}}} & \multicolumn{3}{c|}{\textbf{Electronics}}        \\ \cline{2-7} 
\multirow{-2}{*}{{\color[HTML]{000000} \textbf{Model}}} & R-1                 & R-2                 & R-L                 & R-1            & R-2            & R-L            \\ \hline
\textbf{Seq2Seq}                                        & 19.33               & 9.31                & 18.25               & 22.71          & 11.49          & 21.14          \\ \hline
\textbf{PGNet}                                          & 25.70               & 12.36               & 24.02               & 28.29          & 14.35          & 26.38          \\ \hline
\textbf{STD}                                            & 25.54               & 12.42               & 23.92               & 27.58          & 14.06          & 26.09          \\ \hline
\textbf{HTD}                                            & 27.59               & 12.74               & 25.64               & 28.63          & 14.70          & 27.21          \\ \hline
\textbf{RHTD}                                           & \textbf{28.67}      & \textbf{13.26}      & \textbf{26.58}      & \textbf{31.97} & \textbf{15.23} & \textbf{30.11} \\ \hline
\end{tabular}
\setlength{\abovecaptionskip}{-0.3cm}
\setlength{\belowcaptionskip}{-0.0em}
\caption{\footnotesize ROUGE F1 scores of different methods on two domains' test sets for abstractive review summarization.} %The test sets consists of two domains: \texttt{Healthcare} \&  \texttt{Electronics}.}

\label{table1}
\end{table}

\begin{figure*}
  \begin{minipage}[b]{0.32\textwidth}
    \includegraphics[width=2.5in]{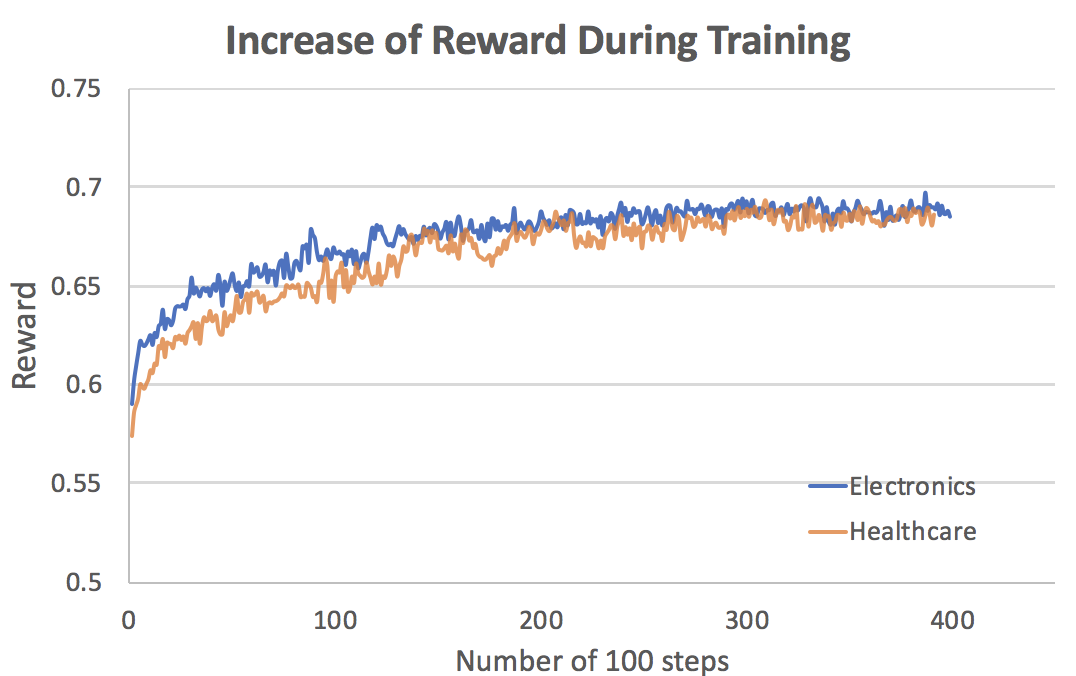}
    \setlength{\abovecaptionskip}{0.0cm}
    \setlength{\belowcaptionskip}{-0.4cm}
    \caption{Learning Curve of \emph{RHTD}}
    \label{fig:reward}
  \end{minipage}
  \hfill
  \begin{minipage}[b]{0.32\textwidth}
    \hbox{\hspace{1.8em}\includegraphics[width=1.78in]{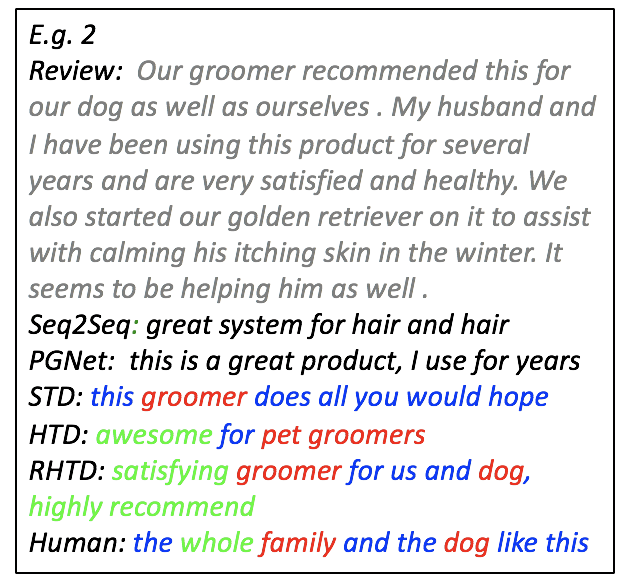}}%
    \setlength{\abovecaptionskip}{0.0cm}
    \setlength{\belowcaptionskip}{-0.4cm}
    \caption{Generated Summaries on E.g. 2}
    \label{fig:eg2}
  \end{minipage}
    \hfill
  \begin{minipage}[b]{0.34\textwidth}
    \hbox{\hspace{0.3em}\includegraphics[width=2.3in]{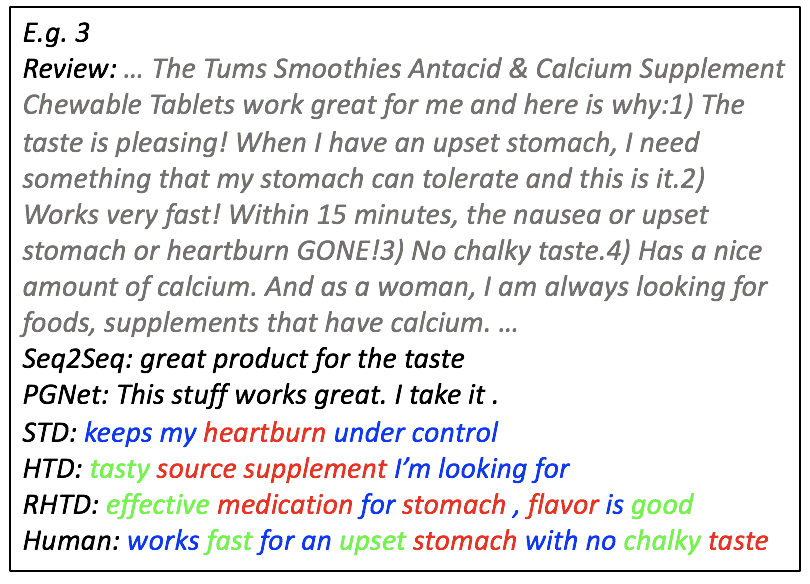}}%
    \setlength{\abovecaptionskip}{0.0cm}
    \setlength{\belowcaptionskip}{-0.4cm}
    \caption{Generated Summaries on E.g. 3}
    \label{fig:eg3}
  \end{minipage}
\end{figure*}

In this subsection, we compare our proposed \emph{RHTD} with the following four strong baseline approaches: 
1). \textbf{\textit{Seq2Seq}}: the standard encoder-decoder RNNs coupled with attention mechanism proposed by \citet{nallapati2016abstractive};
2). \textbf{\textit{PGNet}}: the Pointer-Generator Network proposed by \citet{see2017get} that can both copy words from the source text via pointer network, and produce novel words via the generator;
3). \textbf{\textit{STD}}: the Soft Typed Decoder proposed by \citet{wang2018learning}, which also incorporates three separate decoders for each word type, but simply forces the three decoders to share the whole vocabulary, and employs the weighted sum of the three word distribution as the final word distribution. Note that we adapt the original \textit{STD} model to \textit{PGNet}.
%Note that the original \textit{STD} model used in \cite{wang2018learning} is based on \textit{Seq2Seq}, and we adapt it to \textit{PGNet} by incorporating three additional pointer networks to generate three type-specific copy distributions;
4). \textbf{\textit{HTD}}: our modified Hard Typed Decoder based upon \textit{PGNet}, as introduced in Section~\ref{PA}.1;
5). \textbf{\textit{RHTD}}: our full model with reinforcement learning, as introduced in Section~\ref{PA}.2.

Based on the ROUGE scores reported in Table ~\ref{table1}, it is easy to observe that the performance of \emph{Seq2Seq} is relatively limited.
\emph{PGNet} and \emph{STD} can bring significant improvements over \emph{Seq2Seq} perhaps due to the incorporation of copy mechanism.Moreover, by explicitly incorporating three type-specific decoders, our \emph{HTD} and \emph{RHTD} models can further boost the performance of \emph{PGNet} and \emph{STD} with a large margin. Finally, we can find that \emph{RHTD} consistently outperforms all its competitors, and obtains 3.67\% and 10.66\% performance gains over the second best model for \texttt{Healthcare} and \texttt{Electronics}, respectively. The higher performance on \texttt{Electronics} might result from a larger size of training data.

%, but the margin with the best baseline, \textit{HTD} is 2 percent higher in \textit{Electronics}, because of much more training data. 

\subsection{Further Analysis}
\textbf{Visualization of Rewards Increase: }
To show the advantages of our \emph{RHTD} model, we further plot the reward factor $v_t$ (defined in Eq.\ref{eq:reward_factor}) for both datasets in Fig.~\ref{fig:reward}. Compared to \emph{HTD} (i.e., at step 0), \emph{RHTD} gradually increased the reward by 0.1, meaning that it is better at predicting the right word types (i.e., reference types).

\textbf{Case Study:}
To have a better understanding of the advantage of our model, we select two representative examples in Fig.~\ref{fig:eg2} and Fig.~\ref{fig:eg3} to perform human analysis.%, i.e., the summaries of a multifunctional grooming system manufactured by Philips, and the summaries of a bottle of antacid chewable tablets for pain relief.

First, we can easily observe that all typed decoders indeed generate more informative responses. 
As we can see, \emph{Seq2Seq} tends to generate low-quality summaries like \emph{`great product'} or \emph{`hair and hair'}, and \emph{PGNet} also outputs universal phrases like \emph{`this is a great product'}, or \emph{`I take it'}. On the contrary, all three typed decoders output aspect and opinion words like \emph{`groomer'}, \emph{`pet'}, \emph{`heartburn'}, \emph{`supplement'}, \emph{`medication'} and \emph{`stomach'}, making the summaries more instructive to potential buyers.

Second, \emph{RHTD} is better at extracting the most salient points from input. 
In Fig.~\ref{fig:eg2}, we can see from the original review that this groomer is shared by the purchaser's family and dog. 
While \emph{HTD} only captures partial information and concludes \emph{`awesome for pet groomers'}, \emph{RHTD} gets a more holistic view by outputting \emph{`satisfying groomer for us and dog'}, which is closer to human-uttered summary. 
On the other hand, the review in Fig.~\ref{fig:eg3} covers 4 points regarding the antacid tablet: 1). it is used for stomachache; 2). it works fast; 3). it has no chalky taste; and 4). it contains calcium. 
\emph{Seq2Seq} and \emph{PGNet} generate vague summaries that miss the point.
\emph{STD} is better than the previous two by covering the tablet's usage and effectiveness, but unfortunately copies the wrong word \emph{`heartburn'}.
\emph{HTD} adds that the tablets are \emph{`tasty'}, but mentions nothing about what the medicine is used for. 
Comparatively, \emph{RHTD} is most comprehensive by both stating that it is \emph{`effective'} and that \emph{`flavor is good'}. 
Moreover, \emph{RHTD} is the only non-human model that correctly states the usage: \emph{`medication for stomach'}. 
In summary, \emph{RHTD} takes a more comprehensive look at longer product reviews.
\section{Conclusion}
We presented a two-stage reinforcement learning approach for abstractive review summarization, which first predicts the output word type, and then generates the final word distribution based on the predicted type.
Evaluations on two Amazon product review datasets show the effectiveness of our method. Finally, we believe that the idea of typed decoders can be applied to a variety of NLP tasks.

\section*{Acknowledgments}

This research is supported by the National Research Foundation, Prime Minister's Office, Singapore under its International Research Centres in Singapore Funding Initiative.

\bibliographystyle{ACM-Reference-Format}
\bibliography{sample-base}
\end{document}